\documentclass[letterpaper, 10 pt, conference]{ieeeconf}  

\IEEEoverridecommandlockouts                              

\usepackage[noadjust]{cite}
\usepackage{etoolbox}

\usepackage{amsmath}
\usepackage{amsfonts}
\usepackage{algorithm}
\usepackage{algpseudocode}
\usepackage{url}
\usepackage{multicol}
\usepackage[bookmarks=true]{hyperref}
\usepackage[most]{tcolorbox}
\usepackage{xcolor} 
\usepackage{caption}
\newcommand{\xxnote}[3]{}
\ifx\hidenotes\undefined
  \renewcommand{\xxnote}[3]{\color{#2}{#1: #3}}
\fi

\definecolor{real}{RGB}{242, 140, 7} 
\definecolor{sim}{RGB}{7, 179, 242} 

\makeatletter
\def\blfootnote{\xdef\@thefnmark{}\@footnotetext}
\makeatother

\title{\LARGE \bf
REPeat: A Real2Sim2Real Approach for Pre-acquisition \\ of Soft Food Items in Robot-assisted Feeding

}

\author{Nayoung Ha*$^{1}$, Ruolin Ye*$^{1}$, Ziang Liu$^{1}$, Shubhangi Sinha$^{1}$, Tapomayukh Bhattacharjee$^{1}$
}

\begin{document}

\twocolumn[{%
\renewcommand\twocolumn[1][]{#1}%

\maketitle
\thispagestyle{empty}
\pagestyle{empty}

\vspace{-10pt}
\begin{center}
    \captionsetup{type=figure}
    \includegraphics[width=0.95\textwidth]{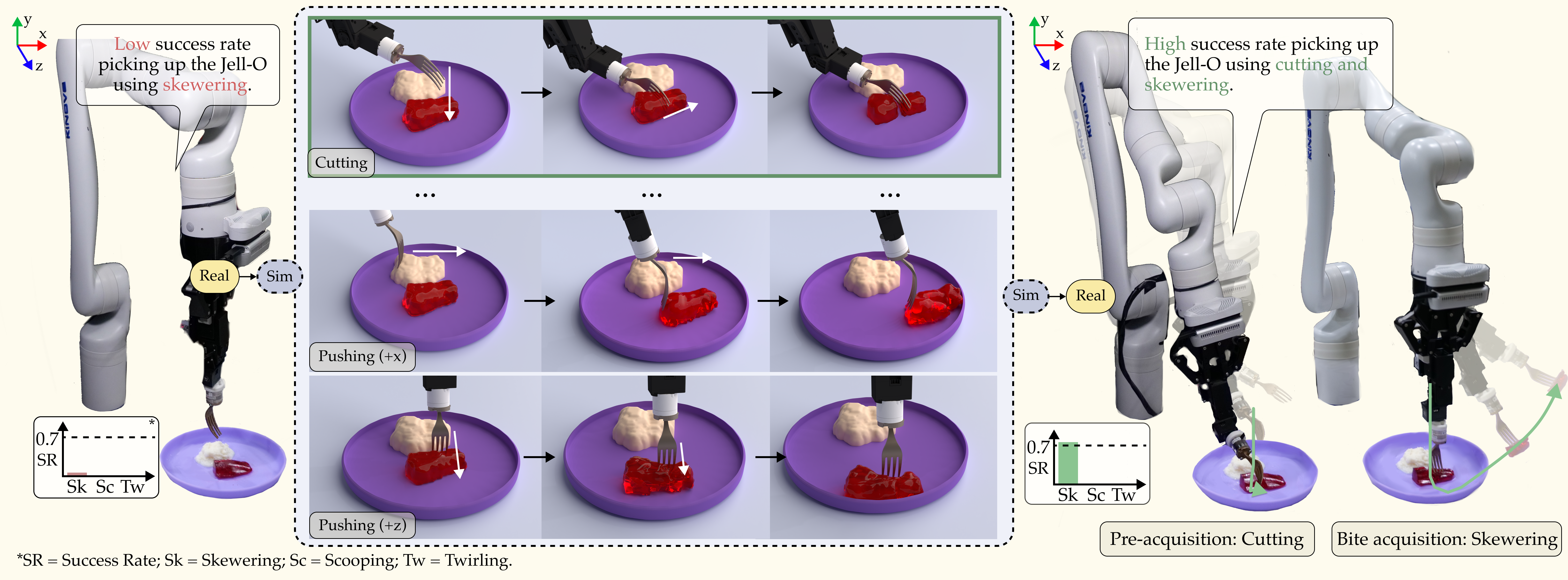}
    \captionof{figure} {We propose \textbf{REPeat}, a Real2Sim2Real system for pre-acquisition of soft food items. The system evaluates the likelihood of successful bite acquisition; if low, it replicates the setup in simulation to explore various pre-acquisition actions. If a certain pre-acquisition action improves the bite acquisition success rate, the robot executes the pre-acquisition and bite acquisition actions in the real world.}
    \label{fig:front}
\end{center}
\vspace{-0.5em}
}]

\blfootnote{\footnotesize *The two authors contributed equally to the paper.}
\blfootnote{\footnotesize $^{1}$Computer Science Department, Cornell University  {\tt\small \{nh285, ry273, zl873, 
ss3392, tapomayukh\}@cornell.edu}}%
\blfootnote{\footnotesize This work was partly funded by National Science Foundation IIS \#2132846, CAREER \#2238792, and DARPA under Contract HR001120C0107. We thank Tom Silver and Rishabh Madan for their feedback.}%

\begin{abstract}
The paper presents REPeat, a Real2Sim2Real framework designed to enhance bite acquisition in robot-assisted feeding for soft foods. It uses `pre-acquisition actions' such as pushing, cutting, and flipping to improve the success rate of bite acquisition actions such as skewering, scooping, and twirling. If the data-driven model predicts low success for direct bite acquisition, the system initiates a Real2Sim phase, reconstructing the food's geometry in a simulation. The robot explores various pre-acquisition actions in the simulation, then a Sim2Real
step renders a photorealistic image to reassess success rates.
If the success improves, the robot applies the action in reality. 
We evaluate the system on 15 diverse plates with
10 types of food items for a soft food diet, showing improvement in bite acquisition success rates by 27\% on average across all plates.
See our project website at \href{https://emprise.cs.cornell.edu/repeat}{emprise.cs.cornell.edu/repeat}. 
\end{abstract}

\vspace{-0.8em}
\section{Introduction}
As of 2021, the World Health Organization \cite{WHO2022GlobalHealth} reports that approximately 1.3 billion individuals, or 16\% of the global population, live with a significant disability. 
Approximately 142 million of these individuals experience \emph{severe} disabilities~\cite{WHO2022GlobalHealth} that limit their independence in performing basic activities of daily living (ADLs).
Among these ADLs, \emph{eating} is especially critical. Robot-assisted feeding systems~\cite{ gordon2020adaptive, feng2019robot, nanavati2023design, bhattacharjee2020moreautonomy,  park2020evaluation, bhattacharjee2019foodhaptics, gordon2021leveraging, sundaresan2023learning, ondras2022human, jenamani2024flair} have the potential to improve the quality of life of care recipients and lessen caregiver workload. 
Previous robot-assisted feeding systems mainly address two subproblems~\cite{madan2022sparcs}: (i) \textit{bite acquisition}~\cite{bhattacharjee2019foodhaptics, gordon2023towards, tai2023scone}, which involves picking up a food item from the plate, and (ii) \textit{bite transfer}~\cite{jenamani2024flair, robotdesign2022kim}, which involves moving the food item near or inside the mouth of a care recipient. In this paper, we focus on bite acquisition. 

We focus on highly deformable soft diet food items for bite acquisition.
A soft diet is essential for care recipients with dysphagia~\cite{jacobsson2000eatingprocess, Healthdirect2022Dysphagia}, a condition characterized by difficulty in swallowing. This  diet includes foods that are easy to chew and swallow~\cite{IDDSI}.
There is a high prevalence of dysphagia with severe mobility limitations, including advanced stages of Amyotrophic Lateral Sclerosis (ALS) (prevalence rate of 80\%) \cite{Onesti2017}, and Parkinson’s disease (80\%) \cite{suttrup2016dysphagia}.

Soft diet food covers a broad spectrum of rheological properties, including Newtonian fluids with various viscosities (e.g., water), non-Newtonian fluids such as Bingham plastics (e.g., mashed potatoes), and pseudoplastics (e.g., oatmeal)~\cite{Gresham2005LearningMA, Zhou2022-zh}, granular solids (e.g., rice)~\cite{GUATEMALA201277}, plastic and elastic solids (e.g., banana, avocado, and Jell-O)~\cite{KALETUNC1991}, and composites of the above (e.g., macaroni and cheese)~\cite{TUNICK2023}. 
Bite acquisition is challenging, especially due to the varying rheological properties of soft foods.

Towards developing a robot-assisted feeding system that can handle soft diet food, we take inspiration from human bite acquisition. Humans often use \emph{pre-acquisition} actions to make acquisition easier~\cite{bhattacharjee2019foodhaptics}. For example, pushing consolidates granular items for enhanced support, cutting adjusts food size for easier skewering, and flipping secures a stable surface to prevent the food from rolling during skewering (e.g., flipping a slice of banana on its side to make it rest on the flat surface, preventing it from rolling while skewering). 

Similarly, pre-acquisition actions have the potential to efficiently enhance robot-assisted bite acquisition success when used intelligently. To explore the large space of pre-acquisition and acquisition actions, one option is to use simulations. Simulation proves particularly valuable for planning and reasoning about irreversible actions, such as cutting food into smaller pieces before performing these actions with an actual robot. Current simulation techniques provide satisfactory models for the dynamics involved in pre-acquisition actions such as moving, tearing, or deforming object~\cite{heiden2021disect}, thus making simulation a useful tool for developing pre-acquisition control policies. However, simulating bite acquisition is still challenging because it requires accurately modeling friction (e.g. friction between a moist, slippery banana and a fork to determine if banana will slip) which is not trivial and hence, the Sim2Real gap for bite acquisition is high~\cite{ipc}.

\textit{Our first key insight is that by leveraging a combination of food simulation for exploring pre-acquisition actions (pushing, cutting, flipping) and a real-world data-driven model for estimating bite acquisition success (e.g., whether the food is on the fork or not), we can effectively mitigate this Sim2Real gap}. While we realize that SOTA food simulations are far from being accurate~\cite{heiden2021disect, Xian2023FluidLabAD} in simulating the exact state of food items after manipulation (e.g., determining the exact shape, size, and texture of a lump of mashed potato after pushing), we realize that these simulations are still reliable enough to estimate final approximate food configuration (e.g., determining if a lump of mashed potato is near the plate wall after pushing). This leads to \textit{our second key insight which suggests that food configuration information after pre-acquisition has enough rich information to inform the success rates of future bite acquisition}.

Using this insight, we propose \textbf{REPeat}, a Real2Sim2Real approach for the pre-acquisition of soft food items. In the REPeat system, a learned module first estimates the success rate of bite acquisition. We build upon SPANet~\cite{feng2019robot} and develop SPANet-soft, an improved food-action prediction model, particularly for food items from a soft diet.  We use this model to decide if it is good enough for direct bite acquisition.
If not, the REPeat system performs Real2Sim to create a set of 3D objects in the simulation.
In this step, 
REPeat uses monocular depth estimation to generate 3D models from real-world RGB images. 
The 3D models are then transferred to a simulation environment~(FluidLab~\cite{Xian2023FluidLabAD}) which then uses the Material Point Method (MPM)~\cite{Sulsky1993APM} to simulate the deformation and tearing effect for the exploration of various pre-acquisition actions. We implemented an adaptive sampling module and a render-on-demand module to enhance FluidLab, enabling it to simulate complex food interactions.
In the Sim2Real step, we use ControlNet~\cite{zhang2023adding}, which utilizes simulated depth data to generate realistic RGB images of the final predicted plate configuration after running the pre-acquisition actions. We then evaluate these generated images to determine the bite acquisition success rate and execute the pre-acquisition action that leads to the state with the highest predicted bite acquisition success rate in the real world. 
Given the lack of high-fidelity food physics simulators, we use an existing MPM-based~\cite{hu2018moving} simulator and design custom simulation environments for soft diet food items. 
We evaluate the proposed approach on 15 diverse food plates with 10 types of soft food items. Our results demonstrate improvement in bite acquisition success rates, underscoring the effectiveness of our approach in advancing feeding assistance for soft diets.

The main contributions of this paper include:
\begin{itemize}
    \item  REPeat, a Real2Sim2Real framework for physics-informed pre-acquisition of soft diets. It leverages monocular depth estimation for 3D modeling (Real2Sim), employs MPM for realistic food physics simulation, and uses ControlNet (Sim2Real) to generate photorealistic RGB images from simulated outcomes for evaluation of pre-acquisition actions. 
    \item High-fidelity simulation environments for food with various rheological properties supporting pre-acquisition actions with large deformations, fractures, or multiphysics coupling effects.
    \item An adaptive particle sampling module and a render-on-demand module, enhancing simulation efficiency and enabling the simulation of complex food interactions with the implementation code released.
    \item SPANet-soft, a network for  prediction of food actions tailored to soft diet items.
    \item Evaluation of our framework on 15 diverse plates with 10 types of soft food items, showing improvement in bite acquisition success rates.
\end{itemize}

\begin{figure*}[ht!]
\centering
\includegraphics[width=0.95\linewidth]{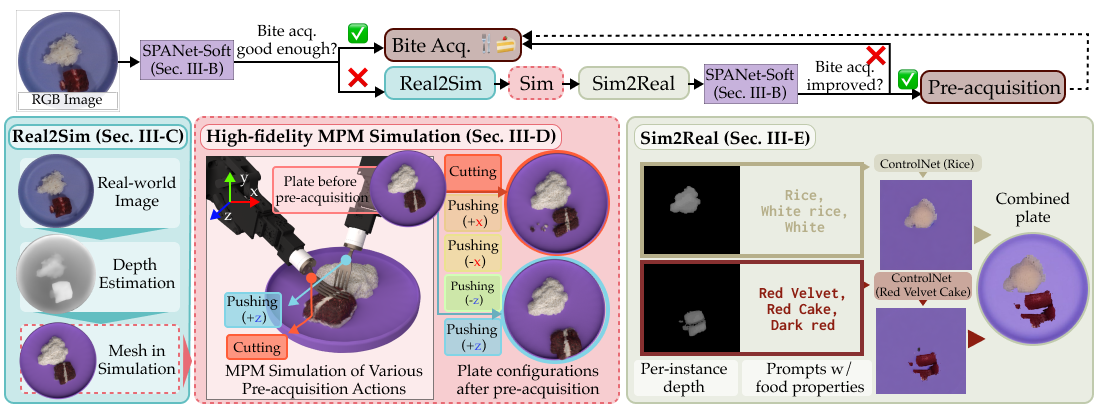}
\vspace{-0.5em}
\caption{\footnotesize \textbf{Overview of REPeat:} The process begins with SPANet-soft (Sec.~\ref{sec:spanetsoft}) giving an initial estimation of the success rate of bite acquisition. The robot performs direct bite acquisition if the initial estimation of the success rate is higher than a threshold. Otherwise, it enters the Real2Sim2Real loop that consists of: (1) \textit{Real2Sim~(Sec.~\ref{sec:real2sim})}: Reconstructing the 3D mesh in real-time with estimated depth as inputs, (2) \textit{Simulation~(Sec.~\ref{sec:simulation})}: Rolling out various pre-acquisition actions using high-fidelity MPM simulation\textcolor{black}{,} (3) \textit{Sim2Real ~(Sec.~\ref{sec:sim2real})}: Rendering a visually realistic picture based on the simulation result.
SPANet-soft \textcolor{black}{evaluates} the result to compare with the success rate of directly picking up food items without pre-acquisition. If the pre-acquisition action improves the bite acquisition success rate, the robot performs the pre-acquisition action first, followed by the bite acquisition action.}
\vspace{-2em}
\label{fig:overview}
\end{figure*}
\vspace{-0.5em}

\section{Related Work}
\subsection{Food Manipulation for Robot-assisted Feeding}
Previous work in robot-assisted feeding has predominantly focused on bite acquisition ~\cite{bhattacharjee2019foodhaptics,  gordon2023towards, tai2023scone} and bite transfer~\cite{jenamani2024bitetransfer, robotdesign2022kim}, with studies exploring techniques for more effective skewering~\cite{feng2019robot, sundaresan2022learning}, twirling~\cite{sundaresan2023learning}, and scooping~\cite{grannen2022learning, tai2023scone, bhaskar2024lava}.  However, research on acquiring foods with different properties such as granular, liquid, and deformable foods~\cite{bhaskar2024lava, AVIL24} has largely been limited to smaller, isolated settings. These studies often fail to capture the complexity of a typical plate setting where food is not isolated, and they lack applicability in caregiving scenarios involving individuals with severe mobility limitations, especially when considering the variety and complexity of soft diet food items.
In contrast, our work evaluates a system across realistic combinations of plates consisting of 15 diverse plates with 10 types of food items, based on soft diet recipes, each displaying distinct rheological properties.

Aside from bite acquisition, existing work has also explored non-acquisition actions such as pushing~\cite{leebite}, cutting~\cite{zhang2019leveraging}, and peeling~\cite{morpheus2023} for food manipulation and meal preparation. Notably, Lee et al. \cite{leebite} employed pushing as a pre-acquisition action for foods such as lettuce leaves and mashed potatoes. FLAIR~\cite{jenamani2024flair} uses the few-shot reasoning capabilities of vision-language foundation models to
plan and execute bite sequences for various pre-acquisition and bite acquisition skills, taking both user preferences and efficiency into consideration. While FLAIR provides the state-of-the-art design of a powerful and modular system, due to the lack of accurate physics information in VLMs, the system can sometimes generate unrealistic outputs. Our system builds on these efforts by incorporating a physics-informed evaluation of multiple pre-acquisition actions, including flipping, pushing, and cutting, to effectively perform bite acquisition on a diverse range of soft diet food items.

\subsection{Simulation for Food Manipulation}
Simulation techniques for food items fall into two main categories: mesh-based and mesh-free methods. Mesh-based methods predominantly utilize the Finite Element Method (FEM) for food modeling. For example, DiSECt~\cite{heiden2021disect} proposes a differentiable FEM-based simulation method to simulate cutting. While these methods offer high physics accuracy, they are computationally intensive due to the need for continuous re-meshing operations. They are also less suited for simulating granular or fluid-like foods. In contrast, mesh-free methods allow more flexibility in deformation modeling, making them the preferred option for simulating fluid-like foods. For instance, FluidLab~\cite{Xian2023FluidLabAD} uses the Material Point Method (MPM) to simulate ice cream, milk, and coffee. MPM has also been used to simulate softer solid foods such as dough~\cite{lin2022diffskill}, given its ability to capture multiphysics coupling, fracture modeling, and large-deformation simulation\textcolor{black}{. This makes it} ideal for simulating a variety of soft diet foods with distinct rheological properties. \textcolor{black}{While RCareWorld 1.0~\cite{RCareWorld}, our own developed simulation platform, supports softbody simulation, it does not support MPM simulation yet.}
Thus, we build our simulation platform upon FluidLab, which allows us to use MPM~\cite{hu2018moving} as the physics backend.

\subsection{Dynamics Modeling for Object Manipulation}
Determining the most effective pre-acquisition action requires understanding how each action affects the state of the food. Previous research has explored learning forward predictive dynamics models from interaction data, predicting future system states from current state and action, using different representations such as pixels~\cite{allenblanchette2020lagnetvip}, particles~\cite{li2020visual}, keypoints~\cite{Manuelli2020KeypointsIT}, and latent variables~\cite{ng2023takes}. Although data-driven dynamics models seem sufficient over a short horizon, they typically suffer from error accumulation in long-horizon prediction, and require a substantial volume of physical interaction data for training. Collecting such data in the real world is time-consuming and wastes a lot of food. In addition, it is difficult to revert the food items to their original states once they have been damaged. Physics-based simulation~\cite{howell2022predictive} is a viable alternative that provides high-fidelity dynamics modeling. Considering the limitations with real-world food items and the potential need for multiple pre-acquisition steps, we model soft-diet food in a physics-based simulator (MPM), to capture their dynamics.


\section{REPeat: Real2Sim2Real Approach for Pre-acquisition of Soft Food Items}
REPeat leverages a Real2Sim2Real approach for selecting pre-acquisition actions (Fig.~\ref{fig:overview}). It takes in an RGB image of the food items on the plate, and predicts a pre-acquisition or bite acquisition action as the output for the robot to take.
The system begins by using SPANet-soft (Sec. \ref{sec:spanetsoft}), a data-driven bite acquisition success rate estimation module, to determine whether direct bite acquisition is good enough for a particular piece of food. If direct bite acquisition is predicted to be challenging, the system transitions to a Real2Sim step~\ref{sec:real2sim}, creating a simulated environment to replicate the food items on the plate. In simulation~\ref{sec:simulation}, the system explores various pre-acquisition actions by executing each action once. Following these actions, it performs a Sim2Real step~\ref{sec:sim2real} to render a photorealistic image for SPANet-soft to estimate the success rate of the bite acquisition actions. After this, the robot executes the pre-acquisition action that leads to the most significant increase in the bite acquisition success rate. We next describe each of these modules in detail (see also the supplementary materials on our website~\cite{repeat24}).

\vspace{-0.5em}
\subsection{Action Space: Pre-acquisition and Bite Acquisition}
In the REPeat system, we consider two types of actions: pre-acquisition actions and bite acquisition actions. We derive the actions marked with `*' from the FLAIR system~\cite{jenamani2024flair} and parameterized \textcolor{black}{them} in an identical way. We give a high-level overview of the actions here, and provide further details of the parameterizations for each of these skills on our website.

The pre-acquisition actions include: 
\begin{itemize}
    \item \underline{Pushing}: We define pushing as a linear motion along the positive or negative direction of the x or z axis, resulting in 4 pushing actions. The pushing action terminates when the food touches other food items or the wall of the plate.
    \item \underline{Cutting*}: We bring the fork horizontal, and rotate the tine to be on its side. We execute a swift downward trajectory to cut the food item, \textcolor{black}{and then a quick flick to separate the two pieces of food items.}
    \item \underline{Flipping}: We bring the tines to the side of the food item, parallel to the major axis. Then, the fork moves quickly perpendicular to the major axis and slightly upward.
\end{itemize}

\begin{figure}[t!]
\centering
\includegraphics[width=1.0\linewidth]{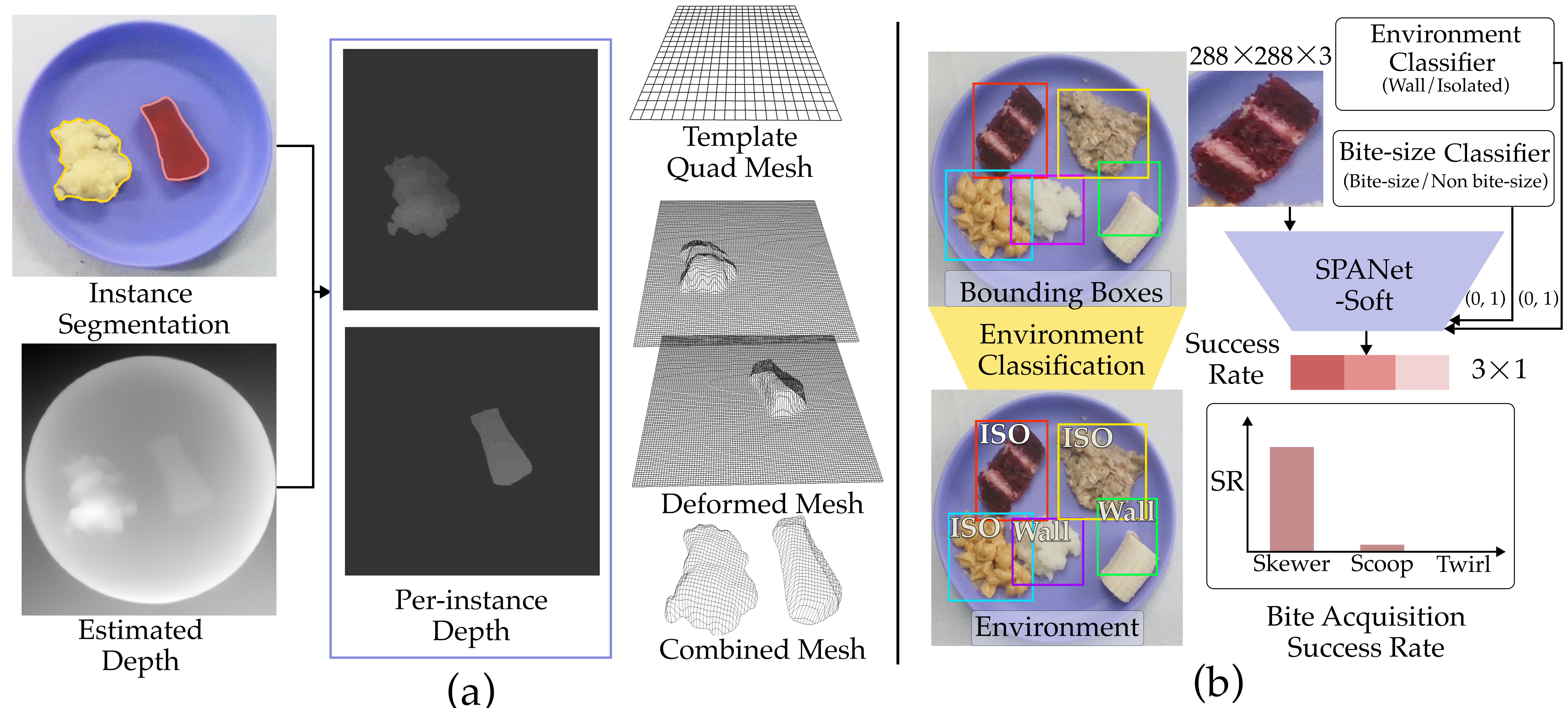}
\caption{\footnotesize
\textbf{(a)} Deformation of the template quad mesh for food mesh reconstruction: Using the RGB image from the camera, we perform instance segmentation, and apply the segmentation mask to the depth map to obtain per-instance depth images. We then use the values of these depth maps as displacement map to deform a template quad mesh. \textbf{(b)} The structure of SPANet-soft.}
\vspace{-2em}
\label{fig:method}
\end{figure}

The bite acquisition actions include:
\begin{itemize}
    \item \underline{Skewering*}: We bring the fork above the center of the food item and then rotate the tines so that the tines are perpendicular to the major axis. We then bring the fork down to let it pierce the food item. After that, the fork performs a scoop-like motion to secure the food item, and \textcolor{black}{brings} it up.
    \item \underline{Scooping*}: We bring tines to a configuration that is horizontal to the plate, and \textcolor{black}{scoop} from the sparsest region to the closest wall support. 
    \item \underline{Twirling*}: We perform twirling on the densest pile of noodles and actuate the roll joint of the fork to perform 2 full twirls.
\end{itemize}

\subsection{SPANet-soft: Food-Action Prediction for Soft Diet Food}
\label{sec:spanetsoft}
Due to the Sim2Real gap in bite acquisition, we \textcolor{black}{develop} the data-driven SPANet-soft module to provide an empirical estimate of the success rate of each bite acquisition action. 

Following the structure of SPANet~\cite{feng2019robot}, SPANet-soft takes in an RGB image of the plate as the input, and predicts the success rate of the bite acquisition actions for each piece of food item as the output. The pipeline has the following components:
\begin{itemize}
    \item \underline{Action Space}: SPANet-soft predicts the success rate for 3 bite acquisition actions, namely skewering, scooping, and twirling derived from FLAIR~\cite{jenamani2024flair}. 
    \item \underline{Food Detection}: We replace RetinaNet~\cite{lin2017focal}, previously used to generate the bounding boxes for the individual food items on the plate, with Grounded-SAM~\cite{ren2024grounded}. 
    \item \underline{Environment Classifier}: We encode the environment surrounding the target food item as a one-hot vector, representing two conditions: (1) Isolated: The target food item is in an empty surrounding. (2) Wall: The target food item is either near the edge of the plate or surrounded by other food pieces.
    \item \underline{Bite-size Classifier}: The classifier estimates the volume of the target food item based on the average height and area extracted from the segmentation mask, providing a one-hot output to indicate whether the item is bite-sized.
\end{itemize}

We show the structure of the pipeline in Fig~\ref{fig:method}(b). The SPANet-soft module takes in a $288\times288\times3$ RGB image cropped and resized using the bounding box of a piece of food item, along with a $2\times 1$ vector for environment classification and a $2\times 1$ vector for bite-size classification. \textcolor{black}{We concatenate these two vectors with the image feature vector from the base network.} It predicts a $3\times 1$ vector containing the success rate for skewering, scooping, and twirling. To train this network, we collect empirical success rates based on real-robot bite acquisition for the 10 types of food items and provide the corresponding images to the network in a way identical to SPANet. \textcolor{black}{We release the dataset on our website~\cite{repeat24}}. We use the smooth L1 loss~\cite{girshick2015fast} between the ground truth vector and the predicted one to ensure the model learns the success rate distribution as closely as possible. Other training details are identical to SPANet. 

\subsection{Real2Sim: Mesh Reconstruction for Food Items}
\label{sec:real2sim}
When SPANet-soft predicts that direct bite acquisition is likely to fail, the Real2Sim2Real pre-manipulation pipeline begins. The Real2Sim step reconstructs a high-quality mesh of the food. This task is challenging since the food items are small and can be subject to noise due to moisture and reflective surfaces. This noise makes it almost impossible to reconstruct the meshes accurately with depth from the sensors. To address this issue, we apply DepthAnything~\cite{depthanything} to perform monocular depth estimation. We reuse the segmentation masks generated by Grounded-SAM~\cite{ren2024grounded} for SPANet-soft and obtain the corresponding depth for each food item. The conventional approaches of reconstructing the mesh with Poisson surface reconstruction~\cite{poisson}, alpha shapes~\cite{edels:3d-alpha}, or ball pivoting~\cite{ballpivoting} are usually slow to create a high-fidelity mesh.
Based on the idea of deforming a template mesh in DepthLab~\cite{du2020depthlab}, we opt for a more computationally efficient method to create the mesh in real time.

Consider $\mathcal{M}$ as a template quadrilateral mesh, consisting of vertices $V=$ $\{p_1, p_2, \ldots, p_n\}$, where each vertex $p_i \in \mathbb{R}^3$ \textcolor{black}{is} represented by $p_i=(x_i, y_i, z_i)$. Given a depth image $D: \mathbb{R}^2 \rightarrow \mathbb{R}$, where $D(u, v)$ specifies the displacement at the corresponding mesh surface point, and assuming mesh $\mathcal{M}$ and depth image $D$ share the same resolution, each pair $(u_i, v_i)$ maps directly to a point $(x_i, y_i)$. We update the position of each vertex by displacing it in the direction of its normal by the depth value, resulting in a new position $v_i^{\prime}$ defined as: $p_i^{\prime}=p_i+D\left(u_i, v_i\right) \cdot n_i$ where $n_i$ is the normal at vertex $p_i$ and $D\left(u_i, v_i\right)$ is the displacement value from the depth map.

This deformation process is applied to every vertex in the mesh, effectively deforming the entire mesh according to the depth information. This leads to an updated set of vertices $V^{\prime}=\{p_1^{\prime}, p_2^{\prime}, \ldots, p_n^{\prime}\}$, which outlines the newly deformed mesh $\mathcal{M}^{\prime}$.

\begin{figure}[t!]
\centering
\includegraphics[width=\linewidth]{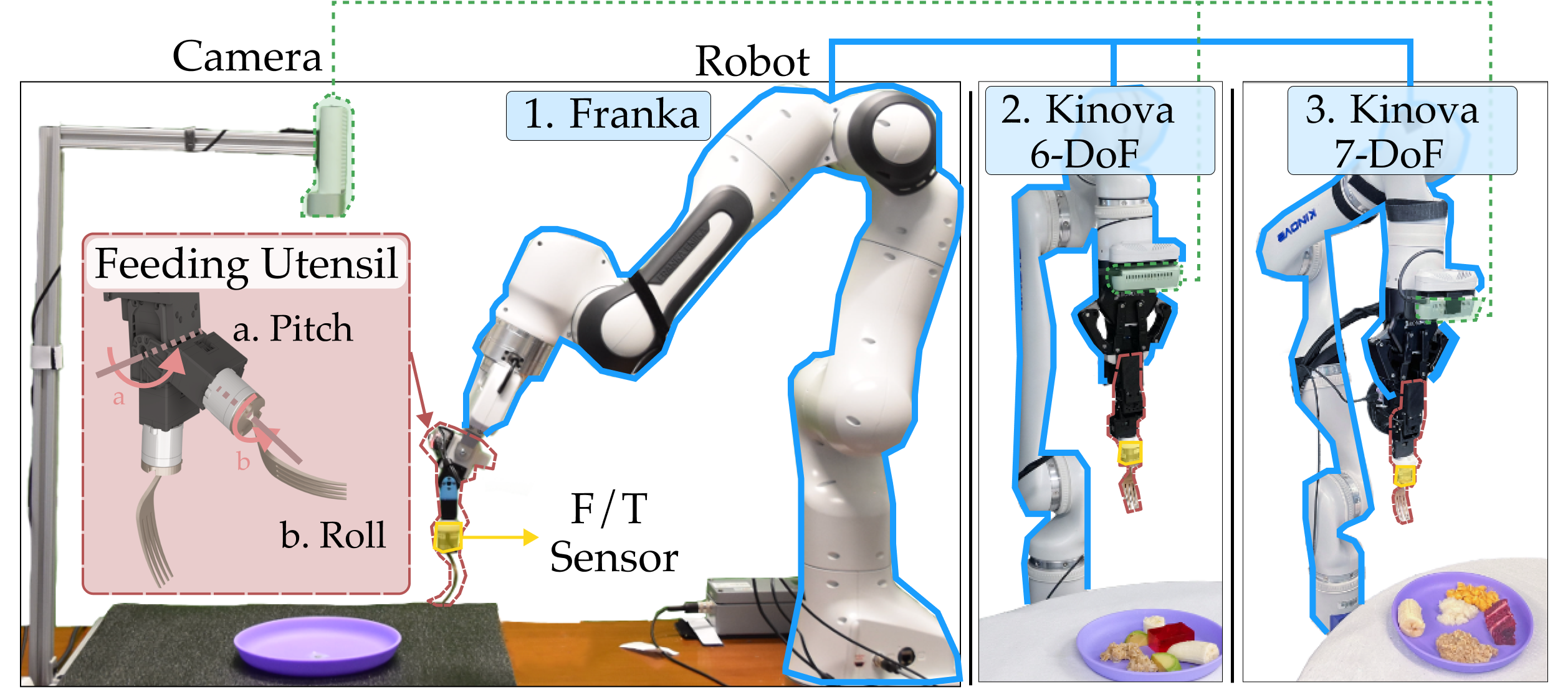}
\vspace{-2em}
\captionsetup{labelfont={color=black},font={color=black}}
\caption{\footnotesize \textcolor{black}{\textbf{Setup:} Our setup features a robot holding a feeding utensil, with a camera for perception and an F/T sensor to detect the end of the pushing action. It is adaptable to various robot embodiments and camera placements (frame or wrist-mounted). The figure shows 3 setups: 1. Franka robot with a camera mounted on a frame 2. Kinova 6-DoF robot with a camera mounted on the wrist 3. Kinova 7-DoF robot with a camera mounted on the wrist. The utensil has 2 DoFs: (a) Pitch, performing a scoop-like motion (b) Roll, performing a twirl-like motion.}}
\label{fig:setup}
\vspace{-2em}
\end{figure}

\subsection{Sim: High-fidelity Simulation for Food Items using MPM}
\label{sec:simulation}
\textcolor{black}{We use the Moving Least Squares Material Point Method (MLS-MPM)~\cite{hu2018moving} to simulate the pre-acquisition actions. MLS-MPM effectively handles complex phenomena such as deformation, fracture, and multi-physics coupling.}
We use the mesh of the food items obtained in the Real2Sim step (detailed in Sec. \ref{sec:real2sim}) as the input. In addition to the food item meshes, \textcolor{black}{the environment includes a fork} that interacts with the food items, and a \textcolor{black}{silicone} plate. We model the food using 3 types of constitutive models, namely the plastic model and elastic model following MLS-MPM~\cite{hu2018moving}, and the elastoplastic model following PlasticineLab~\cite{huang2021plasticinelab}. 
While the actual rheological taxonomy of the food items is more complicated, we approximate them using these 3 models by selecting the closest type of constitutive model for each type of food, with a fixed set of Young's modulus and Lamé constants determined following the parameters in FluidLab~\cite{Xian2023FluidLabAD} detailed on our website~\cite{repeat24}.
We model the fork and the plate as rigid objects using time-varying Signed Distance Fields~(SDFs) created using their mesh files. We simulate the frictional interaction between soft food and rigid objects by calculating the surface normals of the SDFs and applying Coulomb friction~\cite{10.1145/2461912.2461948}. 

\textcolor{black}{FluidLab supports simple interactions with limited objects but cannot simulate the more complex food items on a plate. We implemented 2 modules to enable simulation with multiple objects for a longer horizon. First, we implemented an \emph{adaptive particle sampling} module. It assigns specific densities to each food type instead of using a uniform density for all objects. This allows the simulation of complex food deformations and fractures on a memory-limited GPU. Additionally, we implemented a \emph{render-on-demand} module. It optimizes rendering by generating depth information only after the pre-acquisition action, reducing computational load and enabling long-horizon tasks such as cutting and flipping. These two modules are compatible with FluidLab, enabling it to handle more complex tasks and enhancing its usability. We release this implementation on our project website~\cite{repeat24}.}

\begin{figure*}[t!]
\centering
\includegraphics[width=0.9\linewidth]{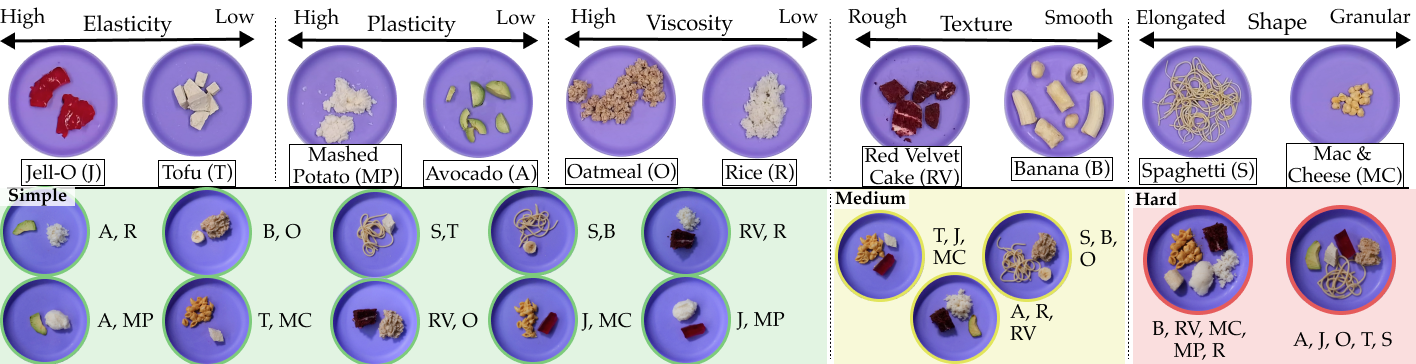}
\caption{\footnotesize \textbf{Upper} 5 axes corresponding to the characteristics of different food items and 10 food types selected to represent the extremes. \textbf{Lower} We evaluate the REPeat system on the following 15 plates containing 10 types of food items. J: Jell-O, MP: Mashed Potato, R: Rice, O: Oatmeal, B: Banana, S: Spaghetti, RV: Red velvet cake, A: Avocado, MC: Mac and cheese, T: Tofu. }
\label{fig:plates}
\vspace{-2em}
\end{figure*}

\subsection{Sim2Real: Rendering Realistic Images from Sim}
\label{sec:sim2real}
After obtaining the \textcolor{black}{predicted} final plate configuration for each pre-acquisition action in simulation, we evaluate the \textcolor{black}{predicted} plate for \textcolor{black}{success} of bite acquisition to select the best pre-acquisition action. To perform this evaluation, we pass the RGB image of the final plate states to SPANet-soft. Although our simulation offers high accuracy in physics, it falls short in visual realism, introducing a Sim2Real gap. We address this gap by generating visually realistic images of the final plate state using the simulated depth data. In particular, we use ControlNet~\cite{zhang2023adding}, a generative model that takes in a control condition and generates an image. Our inputs include depth for the 3D geometry and food category names to generate texture appearance in the pixel domain. We train a category-level ControlNet for each type of food by collecting a \textcolor{black}{dataset} of 
RGB images of the food items on the plate for each category of food items. We also wrote corresponding prompts to provide food properties to the ControlNet. We then create the mesh of the food items in the simulation, and render the depth image of the food item. Using the depth image and prompt as input, we train it to make the generated RGB images as realistic as possible. The network structure, loss function, and other details are identical to the original ControlNet implementation~\cite{zhang2023adding}.
We detail \textcolor{black}{other} data collection and training process of the ControlNet on our website~\cite{repeat24}.
Fig.~\ref{fig:overview} shows the input and example output from ControlNet. 

\section{Evaluation}
We evaluate the system across 15 unique plates each featuring combinations of 10 different types of food to verify the hypothesis that pre-acquisition actions can help improve the success rate of bite acquisition actions for soft diets. 
\textcolor{black}{We evaluate REPeat by comparing it to a baseline that is without pre-acquisition actions.}

\subsection{Experiment Setup}
\paragraph{Hardware}
 We show the real-world hardware setup in Fig. \ref{fig:setup}.  We adopt the utensil proposed by \cite{sundaresan2023learning}. \textcolor{black}{The utensil gives the robot two extra degrees of freedom: the roll and the pitch, and allows precise control of the utensil for actions such as cutting and pushing, while the robot manages the movement between different points in the workspace using Cartesian position control. This feature helps adapt the actions to different robot embodiments.}
 We adapt the utensil to fit the ATI Nano25 F/T sensor for force\textcolor{black}{/torque} sensing.
Our framework is \textcolor{black}{robot-and-camera} setup-agnostic. We \textcolor{black}{evaluated} it using three embodiments: a Kinova Gen3 6-DoF and 7-DoF robot, and a Franka Emika Panda 7-DoF robot. For the two Kinova robots, we use RealSense D435 cameras mounted on their wrists. For the Franka robot, we use an Azure Kinect camera mounted on a fixed frame. We also use a non-slip \textcolor{black}{silicone} plate commonly used in caregiving setups~\cite{feng2019robotassisted} to better simulate the environment of feeding care recipients and prevent the plate from moving.

\paragraph{Simulation}
To simulate soft diet food items, we predefine each food's properties depending on relevant rheological factors. The parameters of the simulation include particle density, elasticity, time step, and shear module, which are calibrated based on food properties and a predetermined set of material properties following Fluidlab~\cite{Xian2023FluidLabAD}. We detail the parameters used for the simulation settings on the project website ~\cite{repeat24}. 

\paragraph{System Details}
The core operations of our system, including data processing activities, are managed by an AMD Ryzen 9 5900X CPU with a base clock speed of 3.7 GHz. For simulation tasks and inference \textcolor{black}{using SAM, DepthAnything, and ControlNet} that require acceleration in graphics rendering, we use an NVIDIA GeForce RTX 3090 GPU with 24 GB GDDR6X memory. We utilize 32 GB DDR4 RAM to support faster operations.

\begin{figure*}[t!]
\centering
\includegraphics[width=0.95\linewidth]{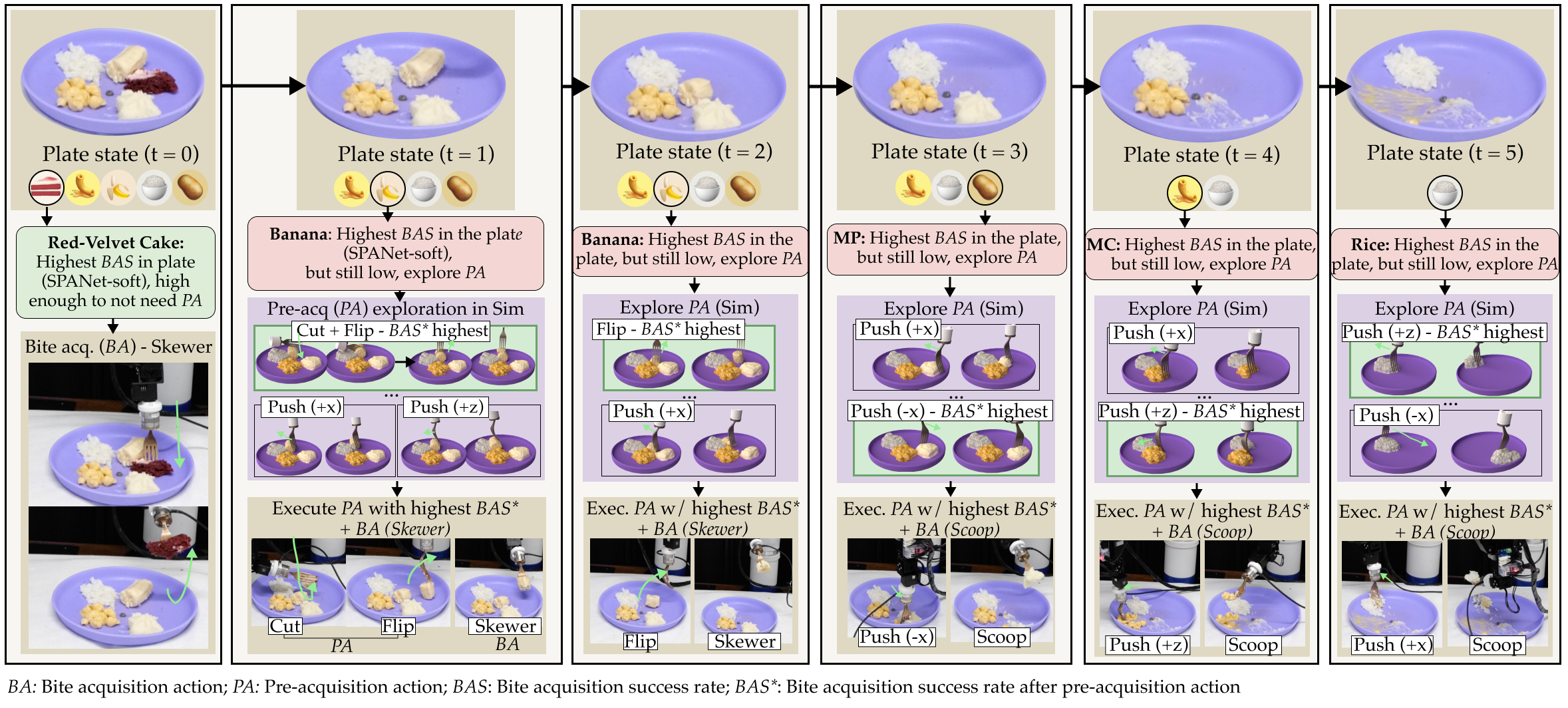}
\vspace{-0.5em}
\caption{\footnotesize An example robot execution sequence for one of the hard plates with five food items. The robot uses pre-acquisition skills such as cutting, flipping, and pushing.}
\label{fig:execute}
\vspace{-1em}
\end{figure*}

\paragraph{Food Selection}
We select 10 types of food representing a diverse range of rheological properties. As illustrated in Fig. \ref{fig:execute} (b), we \textcolor{black}{evaluate} our system with Jell-O, tofu, mashed potato, avocado, oatmeal, rice, red velvet cakes, bananas, spaghetti, and macaroni and cheese~(also called mac \& cheese). These items cover the extremes of the five properties, including elasticity, plasticity, viscosity, texture, and shape, which can affect the bite acquisition success rate. We designed three difficulty levels \textcolor{black}{based on the amount of clutter}: simple, medium, and hard. 
\textcolor{black}{Simple plates have 2 pieces covering 40\%, medium plates 3 pieces covering 60\%, and hard plates 5 pieces covering 80\% of the plate surface area.}

\begin{figure*}[t!]
\centering
\includegraphics[width=0.95\linewidth]{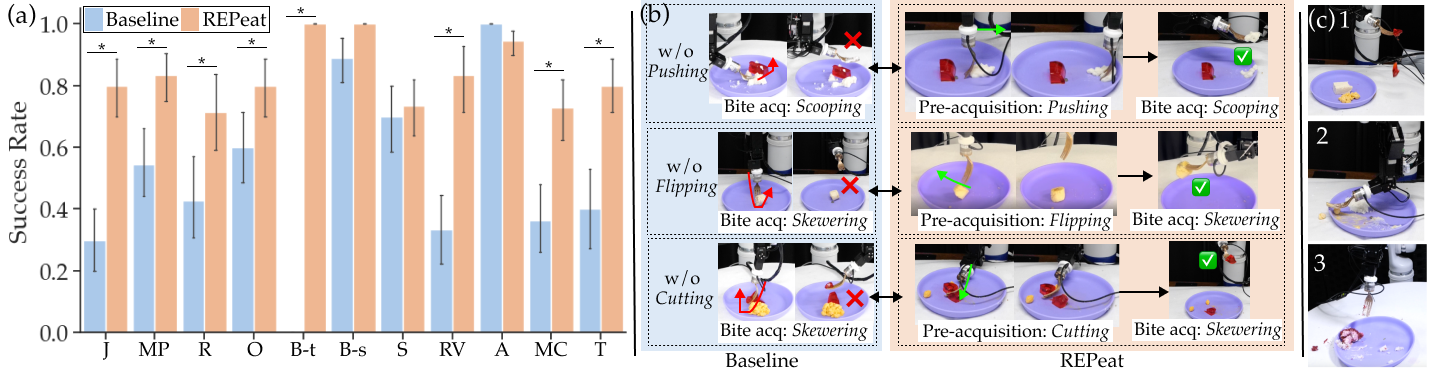}
\caption{\textbf{(a)} Success rate of bite acquisition for the 10 types of food items using the baseline method compared with REPeat. J: Jell-O, MP: Mashed Potato, R: Rice, O: Oatmeal, B-t: Non-bite-sized banana trunk, B-s: Bite-sized banana slice, S: Spaghetti, RV: Red velvet cake, A: Avocado, MC: Mac and cheese, T: Tofu.  \textbf{(b)} Examples of how the pre-acquisition actions help the bite acquisition. \textbf{(c)} Typical failure cases: 1. Fragile food breaks into pieces. 2. Granular food \textcolor{black}{spills} out of the wall of the plate. 3. Multiple food items mixed together and identified as one piece of food item, confusing the perception module. In this case, the white rice is mixed with \textcolor{black}{the white cream} on the red velvet cake. \textcolor{black}{Both the cream and the rice are white, making them very similar. Therefore, rice is confused as the cream, making it identified as an entire piece of food. } }
\label{fig:result}
\vspace{-2em}
\end{figure*}

\subsection{Evaluation Procedure}
We compare our system \textcolor{black}{(w/ pre-acquisition)} with a baseline \textcolor{black}{(w/o pre-acquisition)} to evaluate its effectiveness. \textcolor{black}{The baseline directly uses SPANet-soft to perform the bite acquisition action that leads to the highest success rate.} Based on a bite acquisition success rate set by \textcolor{black}{Gordon} et al. \cite{gordon2023towards}, if the success rate from SPANet-soft is higher than 70\%, we perform a direct bite acquisition. When the success rate is lower, the system performs the pre-acquisition actions \textcolor{black}{that leads to} the highest \textcolor{black}{predicted} bite acquisition success rate. \textcolor{black}{If the pre-acquisition action fails, we repeat it again once. We perform the bite acquisition action after that.}

We use the metric defined in \cite{gordon2023towards} for bite acquisition. \textcolor{black}{
After acquisition, the food must remain on the fork for 3 seconds, the time needed to move it to a care recipient's mouth. Also, we evaluate whether food item is bite-sized using the quantitative metrics from~\cite{gordon2023towards} as the minimum threshold and those from~\cite{jenamani2024flair} as the maximum threshold.}

\subsection{Result}
We present the category-level success rate comparison between our method and the baseline in Fig.~\ref{fig:result} (a). Results indicate that pre-acquisition actions, on average, enhance the bite acquisition success rate by 27\%. We perform chi-square significant tests on the success rates for each food item. The result suggests REPeat performs significantly better for Jell-O, mashed potato, rice, oatmeal, non-bite-sized banana trunk, red velvet cake, mac and cheese, and tofu with p-value $<$ 0.05. We show an execution example for one of the hard plates with 5 pieces of food items in Fig.~\ref{fig:plates} and show the other examples on our website~\cite{repeat24}. We show some of the typical failure cases in Fig.~\ref{fig:result} (c).

The effectiveness of pre-acquisition actions in improving bite acquisition can be attributed to the following factors (illustrated in Fig.~\ref{fig:result} (b)):

\begin{itemize}
\item Pushing consolidates granular food such as mashed potatoes, rice, and mac\&cheese, enhancing scooping success by preventing slippage from the fork. Also, it helps to move the food items near a wall (the wall of the plate or other food items), preventing the food from slipping away.
\item Flipping exposes flat surfaces, crucial for successful skewering actions, as seen with banana slices, where flipping prevents the slice on the side from rolling away during skewering.
\item \textcolor{black}{Cutting fragile food items such as Jell-O into bite-sized pieces helps feeding, and also helps the food items maintain their shape, reducing breakage and the chance of falling off during acquisition.}
\end{itemize}

\section{Discussion}
REPeat takes the first step towards performing physics-informed pre-acquisition actions for a wide variety of soft diet food items. We evaluate the method across 3 different embodiments, 15 combinations of 10 types of food items with various rheological properties. Our evaluation demonstrated that performing physics-informed pre-acquisition actions can increase the success rate of the bite acquisition of soft diet food items.

Through our evaluation, we identify the following limitations that can potentially be resolved and help improve the system in the future: 
\begin{itemize}
\item \underline{Time-varying food properties:} \textcolor{black}{For the food items for soft diets, the rheological properties such as moisture, etc., that affect bite acquisition may vary significantly if placed at room temperature during the course of a meal. In our experiments, we capture food items'  properties for each plate beforehand. However, in real-world feeding, strategies to address time-varying food properties would be beneficial.}
\item \underline{Food perception:} The VLM (GroundedSAM) can detect and segment various food items in our setup. However, due to the visual variety of the food items, we \textcolor{black}{had to carefully construct the prompts}. For example, we specified the red velvet cake as ``red velvet; red brick; red cake; burgundy cake; dark red cake; brown cake; maroon cake; dark purple cake". Advancements in open-set detection and segmentation VLMs might improve the perception pipeline.
\item \underline{Food simulation:} \textcolor{black}{The method we use (MPM)} is computation-heavy and hard to balance between fidelity and speed. \textcolor{black}{We use adaptive sampling and on-demand rendering, making it possible to simulate various food items, but simulating in-the-wild dishes still remain challenging. Future simulation advancements can improve the REPeat system.}
\end{itemize}
Despite these limitations, our system demonstrated \textcolor{black}{that} the use of a Real2Sim2Real framework can help improve the bite acquisition of soft diet food. With future advancements in online food parameter identification, VLMs for perception, and food simulation, we will be able to benefit people with severe mobility limitations who require a soft diet by \textcolor{black}{integrating REPeat with bite-transfer for real-world feeding}.

\bibliographystyle{ieeetr}
\bibliography{references}

\end{document}